\documentclass[letterpaper]{article} 
\usepackage[preprint]{aaai2027}  
\usepackage[hyphens]{url}  
\usepackage{graphicx} 
\usepackage{natbib}  
\usepackage{caption} 
\usepackage{algorithm}
\usepackage{algorithmic}

\usepackage{newfloat}
\usepackage{listings}
\DeclareCaptionStyle{ruled}{labelfont=normalfont,labelsep=colon,strut=off} 
\floatstyle{ruled}
\newfloat{listing}{tb}{lst}{}
\floatname{listing}{Listing}

\usepackage{booktabs}

\usepackage{amsmath}
\usepackage{amssymb}
\usepackage{multirow}
\usepackage[table]{xcolor}
\usepackage{tabularx}
\definecolor{highlight}{RGB}{255,245,210}

\newcommand{\res}[2]{#1\,{\ensuremath{\pm}}\,#2}
\newcommand{\olmoe}{OLMoE}
\newcommand{\qwen}{Qwen}
\newcommand{\deepseek}{DeepSeek}

\newcommand{\Dplus}{\mathcal{D}^{+}}
\newcommand{\Dminus}{\mathcal{D}^{-}}

\usepackage{placeins}
\usepackage{array}
\usepackage{xcolor}
\usepackage{enumitem}
\usepackage{microtype}

\title{TEXAS: Task-Expert-Aware Supervision for Downstream Mixture-of-Experts LLM Adaptation}
\author{
  Guanzhi Deng\textsuperscript{\rm 1},
  Haibo Wang\textsuperscript{\rm 2},
  Kuan Wu\textsuperscript{\rm 1},
  Xiangru Jian\textsuperscript{\rm 3},
  Shing Yin Wong\textsuperscript{\rm 1},\\
  Sichun Luo\textsuperscript{\rm 4},
  Zhuoran Wang\textsuperscript{\rm 1},
  Linqi Song\textsuperscript{\rm 1}\thanks{Corresponding author.}
}
\affiliations{
    \textsuperscript{\rm 1}City University of Hong Kong, Hong Kong, China
    \textsuperscript{\rm 2}Carnegie Mellon University, Pittsburgh, USA \\
    \textsuperscript{\rm 3}University of Waterloo, Ontario, Canada
    \textsuperscript{\rm 4}The University of Hong Kong, Hong Kong, China \\
    \texttt{\url{guanzdeng2-c@my.cityu.edu.hk}, \url{linqi.song@cityu.edu.hk}}
}

\begin{document}

\maketitle

\begin{abstract}
Mixture-of-Experts (MoE) language models route each token through a
small subset of experts, making routing patterns useful for identifying
task-relevant experts during downstream adaptation. Yet current
approaches have two limitations: task experts are typically identified
from aggregate routing statistics that reflect usage rather than
association with successful task completion, and task-expert
activations remain underexplored as signals for supervision allocation. We introduce
\textbf{Task-Expert-Aware Supervision (TEXAS)}, which combines correctness-conditioned task expert discovery
with token-level supervision allocation. TEXAS compares
expert activations on instances that the base model solves successfully
and those it fails to solve, and retains experts more
strongly activated on successful instances. During fine-tuning,
it upweights answer tokens in failed instances when they activate
these experts. TEXAS therefore leverages existing routing
behavior without restricting adaptation to a fixed expert
subset or imposing an explicit target routing distribution. Across three MoE models and six
benchmarks, TEXAS achieves the best or tied-best performance in 17 of
18 settings and improves over the strongest baseline by 1.3--1.5
points on average. Ablations and further analyses validate both the discovered experts and the resulting supervision strategy.
\end{abstract}

\section{Introduction}

Mixture-of-Experts (MoE) architectures scale large language models
(LLMs) by using routers to sparsely activate a subset of experts for
each token, increasing model capacity without proportionally increasing
per-token computation
\cite{shazeer2017,fedus2022switch,lepikhin2021gshard,
jiang2024mixtral}. Beyond computational efficiency, this routing
mechanism gives rise to expert specialization, with different experts
exhibiting distinct functional behaviors and activation patterns across
inputs
\cite{dai-etal-2024-deepseekmoe,xue2024openmoe}. Recent studies further show that routing decisions encode signals that
can be used to identify and leverage task-relevant experts (hereafter
referred to as task experts) for downstream adaptation
\cite{wang2024let,zhou-etal-2024-unveiling,li2025your,
bai-etal-2025-understanding}.

\begin{figure}[!t]
    \centering
    \includegraphics[width=\linewidth]
    {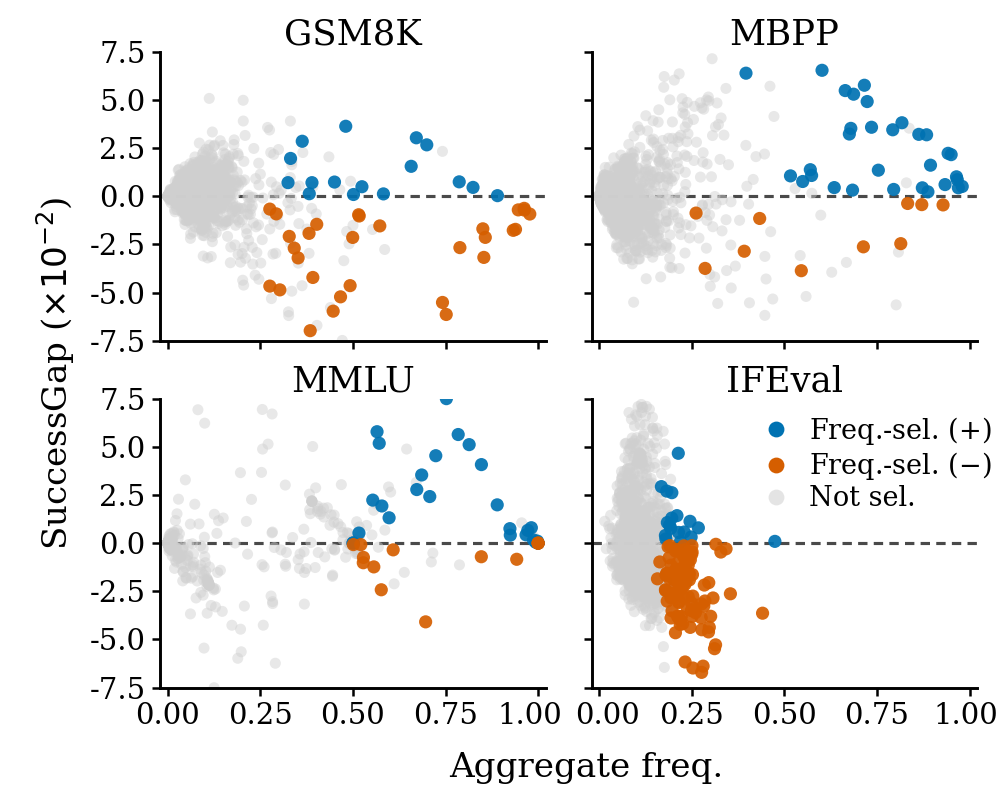}
    \caption{
    Aggregate activation frequency does not reliably identify experts
    associated with successful downstream task completion.
    Each panel uses the corresponding training
    set in Table~\ref{tab:task_setup}, and each point is an OLMoE
    layer--expert pair. The axes show aggregate answer-token activation
    frequency and SuccessGap, defined as the activation-frequency
    difference between base-model successful and failed instances. Blue and orange
    points denote experts selected by ESFT-Token with $p=0.2$~\cite{wang2024let}, separated by positive and
    negative SuccessGap, respectively.
    }
    \label{fig:motivation}
\end{figure}

Existing approaches often identify task experts using routing
statistics aggregated over task data, such as average gate scores or
expert selection frequencies. However, these statistics characterize
aggregate expert usage rather than how expert activations relate to
successful task completion. Figure~\ref{fig:motivation} shows a
consistent mismatch across mathematical reasoning, code generation,
general knowledge, and instruction following: frequency-selected
experts include experts with both positive and negative SuccessGap,
while some unselected experts exhibit substantially larger positive
gaps. Aggregate activation frequency is therefore an incomplete proxy
for identifying experts associated with successful task completion.
Beyond task-expert identification, existing MoE adaptation methods
mainly exploit expert specialization through selective expert
fine-tuning or routing optimization
\cite{wang2024let,bai-etal-2025-understanding,li2026routing,
guo2026advancing}, while using task experts to determine where stronger
training supervision should be allocated remains underexplored.

To address these two limitations, we propose
\textbf{Task-Expert-Aware Supervision (TEXAS)}, a framework for
downstream MoE LLM adaptation. TEXAS first identifies task experts by
comparing their activations between instances that the base model
solves successfully and those it fails to solve. It then uses their
training-time activations to allocate stronger supervision at the token
level. For base-model failed instances, answer tokens that activate the
discovered experts receive higher cross-entropy weights, while standard
supervision is retained elsewhere. TEXAS thus uses task experts to guide supervision allocation, rather
than to select the trainable expert subset or directly prescribe
routing behavior.

We evaluate TEXAS on three MoE LLMs and six downstream benchmarks
covering mathematical reasoning, code generation, general knowledge,
and instruction following. TEXAS achieves the best or tied-best
performance in 17 out of 18 model--task settings and improves average
performance over the strongest baseline by 1.3--1.5 points. Ablation
studies validate the importance of correctness-conditioned expert
discovery, task-expert-aware supervision, and focusing stronger
supervision on base-model failed instances. Further analyses show that
the discovered experts are more strongly associated with successful
task completion and more functionally important than experts selected
by aggregate routing frequency. They also show that TEXAS strengthens
task-expert pathways and directs stronger supervision toward
task-relevant tokens.

Our contributions are summarized as follows:
\begin{itemize}
    \item We identify two limitations in existing approaches that leverage
    task experts for downstream MoE adaptation: aggregate routing
    statistics do not reliably capture experts' association with successful task
    completion, and task experts remain underexplored as signals for
    allocating training supervision.

    \item We introduce TEXAS, which addresses both limitations by discovering task
    experts through correctness-conditioned activation comparisons
    and using their training-time activations to dynamically upweight
    answer tokens that activate them in base-model failed instances.

    \item We conduct extensive experiments across three MoE models
    and six downstream benchmarks. The results demonstrate the
    effectiveness of TEXAS, while controlled ablations and further analyses validate both the discovered
    experts and the resulting supervision allocation.    
\end{itemize}

\section{Related Work}

\paragraph{Task Expert Identification in MoE Models.}
Expert specialization is a central motivation of MoE architectures.
Recent MoE LLMs introduce architectural designs to encourage stronger
specialization, such as fine-grained and shared experts in
DeepSeekMoE~\cite{dai-etal-2024-deepseekmoe}, while analyses of open
MoE models reveal structured expert assignments related to token
identity, input distribution, and model behavior
\cite{xue2024openmoe}. Routing patterns have also been shown to encode
semantic and functional information, serving as training-free
representations~\cite{li2025your} or revealing experts associated with
specific downstream capabilities
\cite{zhou-etal-2024-unveiling,bai-etal-2025-understanding}.
Building on these observations, ESFT identifies task experts using
aggregate statistics such as average gate scores or token selection
ratios~\cite{wang2024let}, whereas CEFT first adapts the router and
then selects context-faithful experts according to their selection
frequency~\cite{bai-etal-2025-understanding}. These methods derive
expert relevance primarily from routing statistics aggregated over
task or capability data. In contrast, TEXAS identifies task experts
by comparing their activations between base-model successful and
failed instances, directly conditioning expert discovery on model
correctness.

\paragraph{Leveraging Expert Specialization for MoE Adaptation.}
Existing methods exploit expert specialization in several ways. ESFT
and CEFT selectively fine-tune the experts identified by their
respective selection procedures. PERFT introduces routed
PEFT modules, while CoMoE uses a contrastive objective between
activated and inactivated experts to promote modularization and
specialization
\cite{liu-etal-2026-parameter,feng-etal-2025-comoe}. Another line of
work modifies routing behavior: RoMA aligns routing weights with those
of semantically similar successful examples~\cite{li2026routing},
R2-T2 performs test-time re-routing toward correctly predicted
neighbors~\cite{li2025rt}, and other methods optimize routing and
specialization through additional objectives or denser router feedback
\cite{guo2026advancing,panda2026dense}. Whereas prior approaches leverage expert specialization through
selective expert updating, specialized adaptation designs, or routing
optimization, TEXAS instead uses task-expert activations to determine
where stronger token-level supervision should be applied.

\section{Method}

\begin{figure*}[t]
    \centering
    \includegraphics[width=0.95\textwidth]{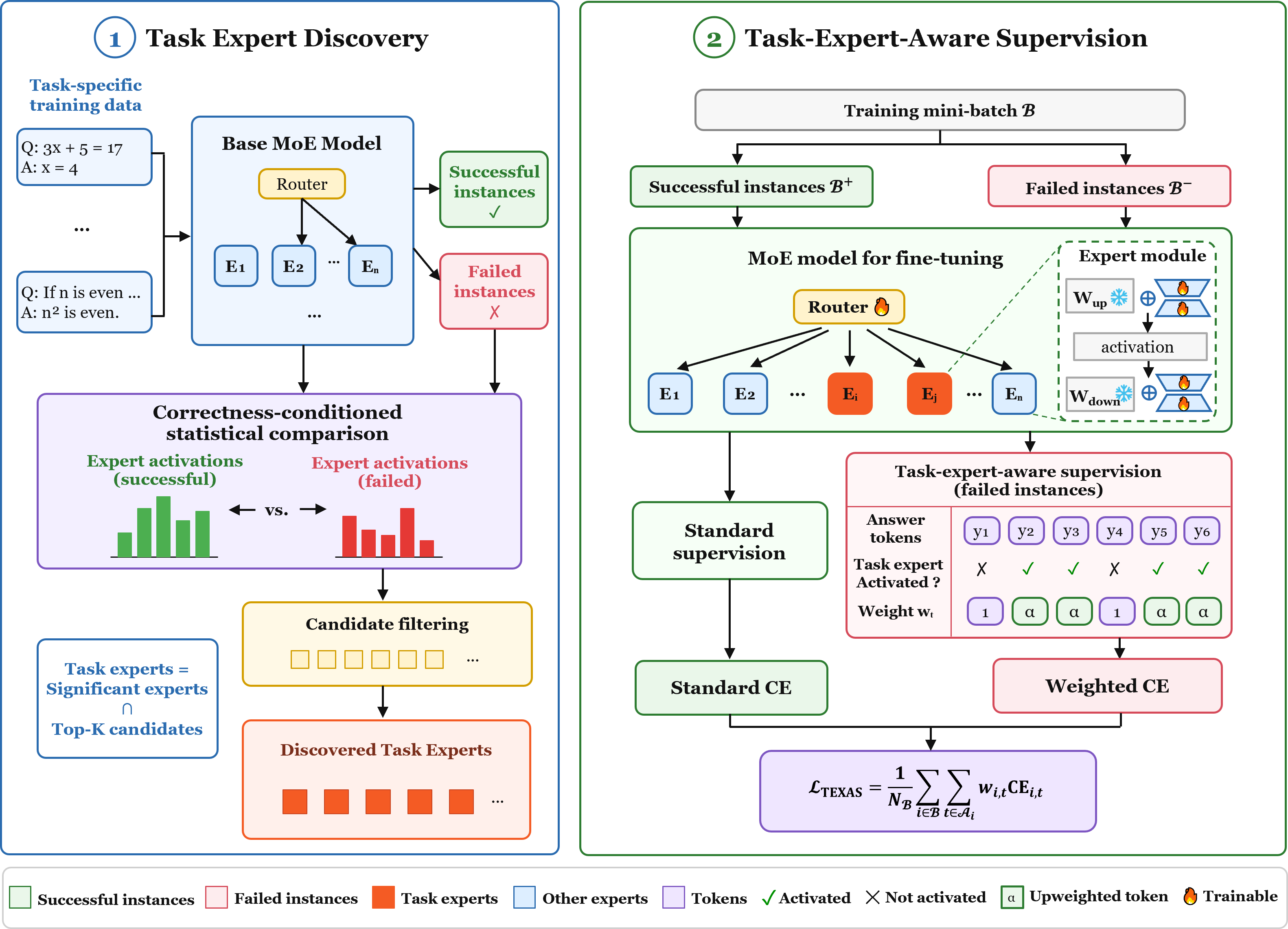}
    \caption{
    Overview of TEXAS. We first split training instances according to base-model success and failure, and identify task experts through correctness-conditioned activation differences with candidate filtering. During fine-tuning, TEXAS increases the supervision weight of answer tokens whose current computation paths activate the discovered task experts.
    }
    \label{fig:framework}
\end{figure*}

TEXAS consists of two tightly coupled components: a correctness-conditioned procedure for discovering task experts, and a task-expert-aware supervision strategy for downstream MoE adaptation. Figure~\ref{fig:framework} gives an overview of the proposed framework.

\subsection{Correctness-Conditioned Task Expert Discovery}

Existing expert-selection methods often identify task experts using aggregate routing statistics, such as activation frequency or gate scores over task data. However, high overall usage does not necessarily indicate that an expert is associated with successful task completion. TEXAS instead discovers task experts by comparing expert activations between instances that the base model solves successfully and those it fails to solve.

Given a downstream training set $\mathcal{D}$ and a base MoE model
$\mathcal{M}_0$, we apply a task-specific inference and evaluation
procedure to each training instance. Instances for which
$\mathcal{M}_0$ satisfies the corresponding success criterion form
the successful subset $\mathcal{D}^{+}$, while all remaining
instances form the failed subset $\mathcal{D}^{-}$. Task-specific inference procedures and success criteria are provided
in Appendix~\ref{app:method_details}.

We then perform teacher-forced forward passes over the reference
answers to collect the model's native top-$k$ routing decisions. For
each instance $i$, layer $\ell$, and expert $e$, we define the
answer-level activation rate as
\[
a_{i,\ell,e}
=
\frac{1}{|\mathcal{A}_i|}
\sum_{t\in\mathcal{A}_i}
\mathbf{1}
\left[e\in R_{i,t}^{(\ell)}\right],
\]
where $\mathcal{A}_i$ denotes the answer-token positions of instance
$i$, and $R_{i,t}^{(\ell)}$ is the set of experts selected by the
model's native top-$k$ router for token $t$ at layer $\ell$.

For each layer-expert pair, we compare its instance-level activation
rates between successful and failed instances using a one-sided
Welch's $t$-test:
\[
H_0:
\mu_{\ell,e}^{+}
\leq
\mu_{\ell,e}^{-},
\qquad
H_1:
\mu_{\ell,e}^{+}
>
\mu_{\ell,e}^{-},
\]
where $\mu_{\ell,e}^{+}$ and $\mu_{\ell,e}^{-}$ denote the mean
answer-level activation rates on $\mathcal{D}^{+}$ and
$\mathcal{D}^{-}$, respectively. We control the false discovery rate
over all layer-expert tests within each model-task setting using the
Benjamini--Hochberg procedure at $q<0.05$. Experts passing the
corrected significance threshold form
$\mathcal{E}_{\mathrm{sig}}^{(\ell)}$.

To exclude statistically significant but rarely activated experts, we
further retain the top-$K$ experts in each layer according to their
mean activation rates on successful instances. The final task-expert
set is
\[
\mathcal{E}_{\mathrm{task}}^{(\ell)}
=
\mathcal{E}_{\mathrm{sig}}^{(\ell)}
\cap
\mathcal{E}_{\mathrm{cand}}^{(\ell)},
\]
where $\mathcal{E}_{\mathrm{cand}}^{(\ell)}$ denotes the resulting
candidate set. Unless otherwise specified, we set $K=2k$, where $k$
is the number of experts selected by the model's native router at each
MoE layer.

\subsection{Task-Expert-Aware Supervision}

After discovering task experts, TEXAS uses their training-time
activations to allocate token-level supervision. The intuition is that,
if an answer token naturally activates experts associated with
successful task completion, this activation provides a signal that the token may lie on a
task-relevant computation path and should receive stronger supervision.

During fine-tuning, TEXAS applies stronger supervision to selected
answer tokens in $\mathcal{D}^{-}$ according to their current routing
decisions. Specifically, we assign the token weight
\[
w_{i,t}
=
\begin{cases}
\alpha, &
i\in\mathcal{D}^{-}
\land
\exists \ell,\,
R_{i,t}^{(\ell)}
\cap
\mathcal{E}_{\mathrm{task}}^{(\ell)}
\neq\emptyset,\\
1, & \text{otherwise},
\end{cases}
\]
where $\alpha>1$ controls the strength of supervision amplification.

For a mini-batch $\mathcal{B}$, let
\[
N_{\mathcal{B}}
=
\sum_{i\in\mathcal{B}}|\mathcal{A}_i|
\]
denote the total number of valid answer tokens. TEXAS optimizes the
mean weighted cross-entropy objective
\[
\mathcal{L}_{\mathrm{TEXAS}}
=
\frac{1}{N_{\mathcal{B}}}
\sum_{i\in\mathcal{B}}
\sum_{t\in\mathcal{A}_i}
w_{i,t}\mathrm{CE}_{i,t}.
\]
Thus, selected answer tokens receive $\alpha$ times their standard
cross-entropy contribution, while all other answer tokens retain unit
weight. When $\alpha=1$, the objective reduces exactly to standard
SFT.

Unlike expert-selection methods such as ESFT~\cite{wang2024let}, TEXAS does not restrict fine-tuning to a fixed subset of experts. Unlike routing-optimization methods such as RoMA~\cite{li2026routing}, TEXAS does not impose a target routing distribution. Instead, it uses naturally occurring task-expert activations as indicators of task-relevant computation paths and strengthens the learning signal at these token positions.

\section{Experiments}

\subsection{Experimental Setup}

We evaluate TEXAS on three MoE LLMs with different architectures and
routing behaviors: \textbf{DeepSeek-V2-Lite}
\citep[][hereafter \deepseek]{liu2024deepseek},
\textbf{OLMoE-1B-7B-0924}
\citep[][\olmoe]{muennighoff2025olmoe}, and
\textbf{Qwen1.5-MoE-A2.7B}
\citep[][\qwen]{qwen_moe}. We consider six downstream tasks spanning
mathematical reasoning, code generation, general knowledge, and
instruction following. For each task, we adapt the model using a
task-specific training set and evaluate it on the corresponding
benchmark. The training sets, evaluation benchmarks, and metrics are
summarized in Table~\ref{tab:task_setup}.

\begin{table}[t]
\centering
\small
\setlength{\tabcolsep}{3pt}
\begin{tabularx}{\columnwidth}{
    @{}
    l
    >{\raggedright\arraybackslash}X
    l
    @{}
}
\toprule
\textbf{Benchmark}
& \textbf{Training Set}
& \textbf{Metric} \\
\midrule
GSM8K
& MetaMathQA-GSM8K~\cite{yu2024metamath}
& Acc. \\

MATH500
& MetaMathQA-MATH
& Acc. \\

HumanEval
& CodeAlpaca~\cite{codealpaca}
& Pass@1 \\

MBPP
& OpenCodeInstruct~\cite{ahmad2025opencodeinstruct}
& Pass@1 \\

MMLU-test
& MMLU-train~\cite{hendrycks2021measuring}
& Acc. \\

IFEval
& RECAST-30K~\cite{guo2026recast}
& Loose Acc. \\
\bottomrule
\end{tabularx}
\caption{
Training sets, evaluation benchmarks, and metrics used in our
experiments. All benchmarks are evaluated in the zero-shot setting.
Loose Acc.\ denotes prompt-level loose accuracy.
}
\label{tab:task_setup}
\end{table}

We compare TEXAS with four baselines. \textbf{Base} denotes the
original model without downstream adaptation, while \textbf{SFT}
denotes standard LoRA-based supervised fine-tuning \cite{hu2022lora}. \textbf{ESFT}
denotes ESFT-Token with $p=0.2$, selectively fine-tuning the
selected experts, whereas \textbf{RoMA} augments the same adaptation
backbone with a routing-alignment objective based on successful
neighboring examples. All fine-tuning methods use a common
LoRA-based expert-adaptation backbone: SFT, RoMA, and TEXAS adapt all
experts, whereas ESFT adapts only the expert subset selected by its
criterion. Router parameters remain trainable for all methods, while
all other pretrained parameters are frozen. Complete implementation details and hyperparameter settings are
provided in Appendix~\ref{app:experimental_setup}, while computational cost analyses are reported
in Appendix~\ref{app:compute}.

\subsection{Main Results}

Table~\ref{tab:main_results} reports the main results across three MoE models and six downstream tasks. TEXAS achieves the best or tied-best performance in 17 out of 18 model-task settings. On average, TEXAS improves over the strongest baseline by 1.5, 1.5, and 1.3 points on \deepseek, \olmoe, and \qwen, respectively. Compared with standard SFT under the same trainable parameter configuration, TEXAS brings average gains of 3.0, 2.4, and 2.7 points on the three models.

The improvements span mathematical reasoning, code generation, and
instruction following. Compared with RoMA, TEXAS improves GSM8K by
2.6, 2.2, and 1.9 points on \deepseek, \olmoe, and \qwen,
respectively, and IFEval by 1.5, 2.2, and 1.9 points. These results suggest that using task-expert activations as fine-grained supervision signals is broadly effective for downstream MoE adaptation. MMLU is the only exception, where TEXAS
performs comparably to the strongest baselines, and we revisit this pattern
in the task-expert pathway analysis in
Figure~\ref{fig:pathway_strengthening}.

\begin{table*}[t]
\centering
\small
\begin{tabular}{lccccccc}
\toprule
\textbf{Method} 
& \textbf{GSM8K} 
& \textbf{MATH500} 
& \textbf{HumanEval} 
& \textbf{MBPP} 
& \textbf{MMLU} 
& \textbf{IFEval}
& \textbf{Avg.} \\
\midrule

\multicolumn{8}{c}{\textit{DeepSeek-V2-Lite}} \\
\midrule
Base  
& \res{5.3}{0.8} 
& \res{6.8}{0.4} 
& \res{25.0}{1.6} 
& \res{25.7}{0.7} 
& \res{46.9}{0.1} 
& \res{14.2}{0.9} 
& \res{20.7}{0.8} \\
SFT   
& \res{57.9}{0.5} 
& \res{16.2}{0.7} 
& \res{27.6}{0.8} 
& \res{31.0}{0.7} 
& \res{56.0}{0.1} 
& \res{26.5}{0.8} 
& \res{35.9}{0.6} \\
ESFT  
& \res{59.1}{0.7} 
& \res{16.6}{0.7} 
& \res{29.3}{1.2} 
& \res{32.5}{0.5} 
& \res{56.2}{0.1} 
& \res{27.8}{0.5} 
& \res{36.9}{0.6} \\
RoMA  
& \underline{59.9}\,{\ensuremath{\pm}}\,\underline{0.8} 
& \underline{17.1}\,{\ensuremath{\pm}}\,\underline{0.5}
& \underline{29.9}\,{\ensuremath{\pm}}\,\underline{0.7}
& \underline{33.1}\,{\ensuremath{\pm}}\,\underline{0.4}
& \textbf{56.3\,{\ensuremath{\pm}}\,0.1} 
& \underline{28.2}\,{\ensuremath{\pm}}\,\underline{0.7} 
& \underline{37.4}\,{\ensuremath{\pm}}\,\underline{0.5} \\
\textbf{TEXAS (Ours)} 
& \textbf{62.5\,{\ensuremath{\pm}}\,1.0} 
& \textbf{18.6\,{\ensuremath{\pm}}\,0.4} 
& \textbf{31.5\,{\ensuremath{\pm}}\,0.9} 
& \textbf{34.7\,{\ensuremath{\pm}}\,0.5} 
& \textbf{56.3\,{\ensuremath{\pm}}\,0.1} 
& \textbf{29.7\,{\ensuremath{\pm}}\,0.8} 
& \textbf{38.9\,{\ensuremath{\pm}}\,0.6} \\

\midrule
\multicolumn{8}{c}{\textit{OLMoE-1B-7B-0924}} \\
\midrule
Base  
& \res{4.7}{0.3} 
& \res{2.8}{0.4} 
& \res{12.8}{0.3} 
& \res{3.9}{0.5} 
& \res{46.2}{0.1} 
& \res{16.8}{0.7} 
& \res{14.5}{0.4} \\
SFT   
& \res{30.8}{0.2} 
& \res{4.9}{0.3} 
& \res{13.8}{0.5} 
& \res{13.8}{0.6} 
& \res{51.0}{0.1} 
& \res{26.9}{0.8} 
& \res{23.5}{0.4} \\
ESFT  
& \res{31.5}{0.4} 
& \res{5.5}{0.5} 
& \res{14.3}{0.4} 
& \res{14.2}{0.3} 
& \textbf{51.3\,{\ensuremath{\pm}}\,0.1} 
& \res{27.8}{0.5} 
& \res{24.1}{0.4} \\
RoMA  
& \underline{32.2}\,{\ensuremath{\pm}}\,\underline{0.6} 
& \underline{5.8}\,{\ensuremath{\pm}}\,\underline{0.4} 
& \underline{14.6}\,{\ensuremath{\pm}}\,\underline{0.6} 
& \underline{14.5}\,{\ensuremath{\pm}}\,\underline{0.5} 
& \res{51.1}{0.1} 
& \underline{28.4}\,{\ensuremath{\pm}}\,\underline{0.8} 
& \underline{24.4}\,{\ensuremath{\pm}}\,\underline{0.5} \\
\textbf{TEXAS (Ours)} 
& \textbf{34.4\,{\ensuremath{\pm}}\,0.5} 
& \textbf{6.9\,{\ensuremath{\pm}}\,0.3} 
& \textbf{15.9\,{\ensuremath{\pm}}\,0.5} 
& \textbf{16.1\,{\ensuremath{\pm}}\,0.2} 
& \underline{51.2}\,{\ensuremath{\pm}}\,\underline{0.1} 
& \textbf{30.6\,{\ensuremath{\pm}}\,0.7} 
& \textbf{25.9\,{\ensuremath{\pm}}\,0.4} \\

\midrule
\multicolumn{8}{c}{\textit{Qwen1.5-MoE-A2.7B}} \\
\midrule
Base  
& \res{23.7}{0.4} 
& \res{11.8}{0.5} 
& \res{33.3}{0.3} 
& \res{22.0}{0.5} 
& \res{60.1}{0.1} 
& \res{21.9}{0.3} 
& \res{28.8}{0.4} \\
SFT   
& \res{65.7}{0.3} 
& \res{22.0}{0.5} 
& \res{40.2}{0.6} 
& \res{28.1}{0.7} 
& \textbf{62.2\,{\ensuremath{\pm}}\,0.1} 
& \res{34.2}{0.1} 
& \res{42.1}{0.4} \\
ESFT  
& \res{66.8}{0.7} 
& \res{23.1}{0.4} 
& \res{41.6}{0.5} 
& \res{28.5}{0.4} 
& \res{62.1}{0.1} 
& \res{35.5}{0.4} 
& \res{42.9}{0.4} \\
RoMA  
& \underline{67.7}\,{\ensuremath{\pm}}\,\underline{0.5} 
& \underline{23.7}\,{\ensuremath{\pm}}\,\underline{0.6} 
& \underline{42.1}\,{\ensuremath{\pm}}\,\underline{0.3} 
& \underline{28.9}\,{\ensuremath{\pm}}\,\underline{0.6} 
& \textbf{62.2\,{\ensuremath{\pm}}\,0.1} 
& \underline{36.4}\,{\ensuremath{\pm}}\,\underline{0.3} 
& \underline{43.5}\,{\ensuremath{\pm}}\,\underline{0.4} \\
\textbf{TEXAS (Ours)} 
& \textbf{69.6\,{\ensuremath{\pm}}\,0.4} 
& \textbf{25.0\,{\ensuremath{\pm}}\,0.2} 
& \textbf{43.3\,{\ensuremath{\pm}}\,0.5} 
& \textbf{30.2\,{\ensuremath{\pm}}\,0.2} 
& \textbf{62.2\,{\ensuremath{\pm}}\,0.1} 
& \textbf{38.3\,{\ensuremath{\pm}}\,0.6} 
& \textbf{44.8\,{\ensuremath{\pm}}\,0.3} \\

\bottomrule
\end{tabular}
\caption{
Main results across three MoE models and six downstream tasks.
Results are reported as mean $\pm$ std over three random seeds.
\textbf{Bold} indicates the best result in each model-task group, and \underline{underline} indicates the second-best distinct result.
}
\label{tab:main_results}
\end{table*}

\subsection{Ablation and Robustness}

\paragraph{Ablation study.}
We compare TEXAS with SFT and four controlled variants on OLMoE using
GSM8K, MBPP, and IFEval. \textbf{TEXAS-Freq} replaces the discovered
task experts with the highest-frequency experts in each layer, matching
the number selected by TEXAS, while retaining the same token-weighting
mechanism. \textbf{TEXAS-Route} retains the same task experts and token
positions selected by TEXAS but replaces loss upweighting with an
auxiliary objective that increases routing mass toward these experts.
The remaining two variants alter only the scope of token upweighting:
\textbf{TEXAS-AllInst} upweights task-expert-activating tokens in all
training instances, whereas \textbf{TEXAS-AllTok} upweights all answer
tokens in base-model failed instances.

\begin{table}[t]
\centering
\small
\setlength{\tabcolsep}{3pt}
\begin{tabular}{@{}lccccc@{}}
\toprule
\textbf{Variant}
& \textbf{GSM8K}
& \textbf{MBPP}
& \textbf{IFEval}
& \textbf{Avg.}
& \textbf{$\Delta$} \\
\midrule

SFT
& \res{30.8}{0.2}
& \res{13.8}{0.6}
& \res{26.9}{0.8}
& \res{23.8}{0.5}
& +0.0 \\

T-Freq
& \res{32.8}{0.6}
& \res{14.6}{0.8}
& \res{27.8}{0.9}
& \res{25.1}{0.8}
& +1.3 \\

T-Route
& \res{31.9}{0.4}
& \res{14.1}{0.6}
& \res{27.4}{0.6}
& \res{24.5}{0.5}
& +0.7 \\

T-AllInst
& \underline{33.0}\,{\ensuremath{\pm}}\,\underline{0.3}
& \underline{15.0}\,{\ensuremath{\pm}}\,\underline{0.5}
& \underline{28.4}\,{\ensuremath{\pm}}\,\underline{0.8}
& \underline{25.5}\,{\ensuremath{\pm}}\,\underline{0.5}
& \underline{+1.7} \\

T-AllTok
& \res{31.8}{0.4}
& \res{14.3}{0.3}
& \res{27.1}{0.5}
& \res{24.4}{0.4}
& +0.6 \\

\textbf{TEXAS}
& \textbf{34.4\,{\ensuremath{\pm}}\,0.5}
& \textbf{16.1\,{\ensuremath{\pm}}\,0.2}
& \textbf{30.6\,{\ensuremath{\pm}}\,0.7}
& \textbf{27.0\,{\ensuremath{\pm}}\,0.5}
& \textbf{+3.2} \\

\bottomrule
\end{tabular}
\caption{
Ablation results on OLMoE. In variant names, ``T'' abbreviates
``TEXAS''. $\Delta$ denotes the average improvement
over SFT.
}
\label{tab:ablation}
\end{table}

As shown in Table~\ref{tab:ablation}, all variants improve over SFT,
but full TEXAS performs best on all three tasks and yields the largest
average gain (+3.2 points). The smaller gains of TEXAS-Freq (+1.3) and
TEXAS-Route (+0.7) support correctness-conditioned discovery over
frequency-based selection and loss upweighting over the auxiliary
routing objective, respectively. TEXAS-AllInst (+1.7) and
TEXAS-AllTok (+0.6) further show that weighting is most effective when
restricted to task-expert-activating tokens in base-model failed
instances.

\paragraph{Hyperparameter robustness.}
TEXAS has two main hyperparameters: the candidate filtering size $K$
and the token-level loss weight $\alpha$. We evaluate OLMoE on GSM8K,
MBPP, and IFEval under
$K\in\{k,1.5k,2k,3k,\text{No filt.}\}$ and
$\alpha\in\{1.05,1.1,1.2,1.5,2.0\}$, where $k$ is the number of
experts activated at each MoE layer. \textit{No filt.} retains all
statistically significant experts, while $\alpha=1$ reduces TEXAS to
standard SFT.

\begin{figure}[t]
    \centering
    \includegraphics[width=\linewidth]{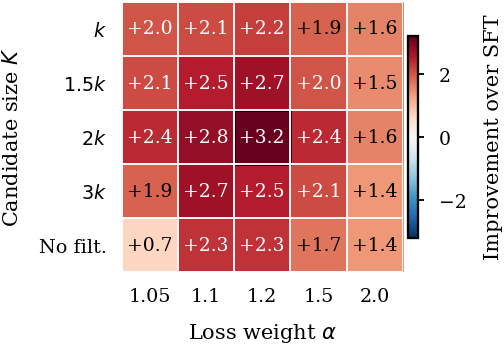}
    \caption{
    Sensitivity of TEXAS to candidate filtering size $K$ and loss
    weight $\alpha$ on OLMoE. Each cell reports the average performance
    improvement over the corresponding SFT baselines across GSM8K,
    MBPP, and IFEval, with the model fine-tuned separately for each
    task.
    }
    \label{fig:hyperparameter}
\end{figure}

As shown in Figure~\ref{fig:hyperparameter}, all tested configurations
outperform SFT, with average gains ranging from $+0.7$ to $+3.2$
points across the three tasks. TEXAS performs best with moderate
candidate sizes ($1.5k$--$3k$) and loss weights ($1.1$--$1.5$),
whereas removing candidate filtering or increasing $\alpha$ to $2.0$
yields smaller improvements. These results suggest that a focused
expert set and moderate supervision amplification are beneficial.
Among the tested settings, $K=2k$ and $\alpha=1.2$ achieve the largest
average gain ($+3.2$) and are used as the default throughout the paper
unless otherwise specified. Additional
experimental details for the ablation and robustness studies are
provided in Appendix~\ref{app:ablation_robustness}.

\section{Why TEXAS Works}

We examine why TEXAS works from three complementary perspectives:
whether correctness-conditioned discovery identifies experts that
remain associated with successful task completion and are functionally
important, whether task-expert-aware supervision reinforces their
computation pathways during adaptation, and whether the upweighted
tokens are enriched for task-relevant content.

\paragraph{TEXAS discovers success-associated, functionally important
experts.} To test whether task experts discovered from training data remain
associated with successful task completion on unseen benchmarks, we
evaluate their SuccessGap on the corresponding benchmark instances.
We compare the correctness-conditioned experts with those identified
by ESFT from aggregate token-routing statistics on the same training
data, together with a random-selection baseline.

For an expert set $\mathcal{E}_{\tau}$ associated with task $\tau$, we
compute
\[
\operatorname{SuccessGap}(\mathcal{E}_{\tau})
=
\frac{1}{|\mathcal{E}_{\tau}|}
\sum_{(\ell,e)\in\mathcal{E}_{\tau}}
\left(
\mu^{+,\mathrm{bench}}_{\ell,e}
-
\mu^{-,\mathrm{bench}}_{\ell,e}
\right),
\]
where $\mu^{+,\mathrm{bench}}_{\ell,e}$ and
$\mu^{-,\mathrm{bench}}_{\ell,e}$ denote the mean answer-level
activation rates of expert $e$ in layer $\ell$ on benchmark instances
that the base model solves successfully and fails to solve,
respectively. A larger SuccessGap indicates that the selected experts
are more active on successfully solved instances than on failed ones.

\begin{table}[t]
\centering
\small
\setlength{\tabcolsep}{1.6pt}
\begin{tabular}{@{}lccccccc@{}}
\toprule
\textbf{Selection}
& \multicolumn{7}{c}{\textbf{SuccessGap ($\times 10^{-2}$) $\uparrow$}} \\
\cmidrule(lr){2-8}
& \textbf{GSM}
& \textbf{MATH}
& \textbf{HE}
& \textbf{MBPP}
& \textbf{MMLU}
& \textbf{IF}
& \textbf{Avg.} \\
\midrule

Random
& -0.004
& -0.005
& -0.019
& 0.003
& 0.004
& -0.007
& -0.005 \\

ESFT
& 1.970
& 1.280
& 2.810
& 1.490
& 0.880
& 4.010
& 2.073 \\

\textbf{CC}
& \textbf{5.990}
& \textbf{4.230}
& \textbf{4.240}
& \textbf{5.910}
& \textbf{1.420}
& \textbf{6.230}
& \textbf{4.670} \\

\bottomrule
\end{tabular}
\caption{
SuccessGap on unseen benchmarks using \olmoe~for randomly selected experts and task
experts identified by ESFT or correctness-conditioned (CC) discovery.
Random reports the mean over 1,000 expert sets matched to the
cardinality of the CC set.
}
\label{tab:expert_success_gap}
\end{table}

As shown in Table~\ref{tab:expert_success_gap},
correctness-conditioned task experts exhibit larger SuccessGap than
those identified by ESFT on all six benchmarks, increasing the average
from $2.07\times10^{-2}$ to $4.67\times10^{-2}$. In contrast, the SuccessGap of randomly sampled
expert sets remains centered near zero across all tasks. These results
show that task experts identified through correctness-conditioned
discovery preserve a stronger association with successful task
completion on unseen benchmarks, relative to both aggregate-routing
and random expert selection.

We next assess whether the discovered task experts are functionally
important by masking them in the SFT-adapted OLMoE and measuring the resulting
performance degradation. We compare correctness-conditioned experts
with experts identified by ESFT and randomly selected experts under the
same layer-wise masking budget. The complete masking protocol is provided in Appendix~D.

As shown in Figure~\ref{fig:masking}, masking the
correctness-conditioned experts causes the largest performance
degradation across all six tasks. This result indicates that these
experts make a stronger functional contribution to downstream task
performance than those identified by ESFT or random selection.
The SuccessGap results on unseen benchmarks and the masking analysis
therefore show that correctness-conditioned discovery identifies
experts that are both associated with successful task completion and
important for downstream task performance.

\begin{figure}[t]
    \centering
    \includegraphics[width=\linewidth]{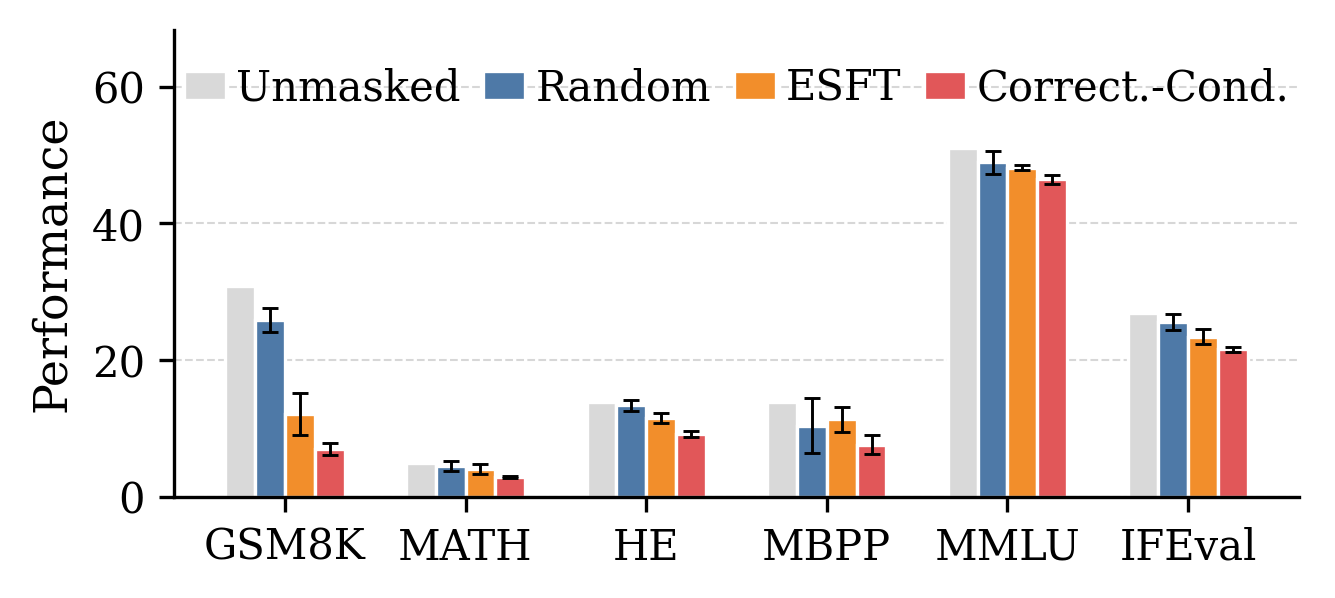}
    \caption{
    Performance of SFT-adapted OLMoE after expert masking. The three conditions
    use the same layer-wise masking budget, with selected experts
    masked by zeroing their down-projection matrices. Error bars denote
    standard deviations over three independently sampled mask sets.
    }
    \label{fig:masking}
\end{figure}

\paragraph{TEXAS reinforces task-expert pathways.} TEXAS does not explicitly impose a routing objective or force tokens
to select the discovered task experts. We therefore examine whether
task-expert-aware supervision can nevertheless strengthen these
computation pathways during adaptation.

\begin{figure}[t]
    \centering
    \includegraphics[width=\linewidth]
    {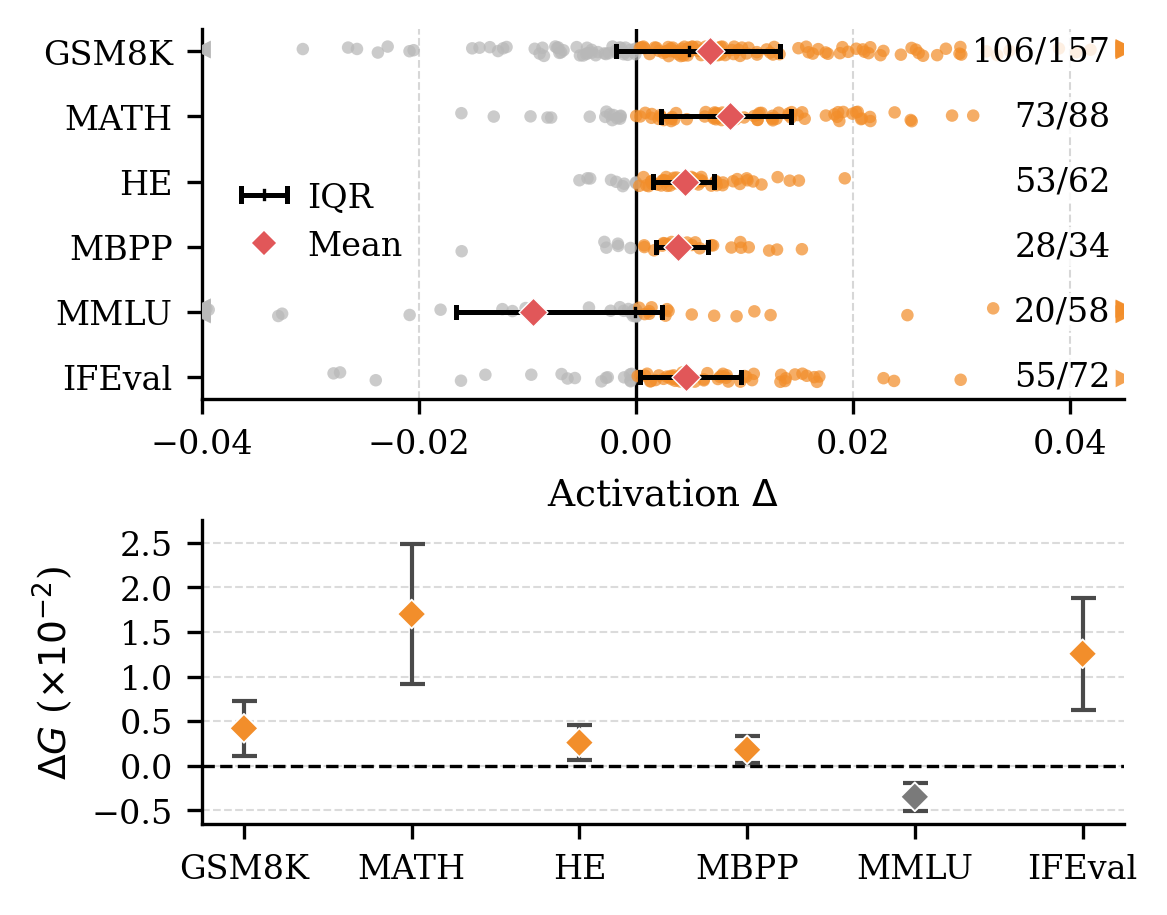}
    \caption{
    Task-expert pathway changes on OLMoE.
    Top: per-expert activation-frequency differences between TEXAS and
    SFT; positive values indicate higher activation under TEXAS.
    Values on the right report the proportion
    of task experts with positive activation changes.
    Bottom: the TEXAS--SFT difference in the correction-conditioned
    activation gap, $\Delta G=G_{\mathrm{TEXAS}}-G_{\mathrm{SFT}}$.
    Positive values favor TEXAS; error bars denote 95\% confidence
    intervals.
    }
    \label{fig:pathway_strengthening}
\end{figure}

At the expert level, the activation-frequency difference between
TEXAS and SFT is computed for each identified task expert on the same
evaluation data. The top panel of Figure~\ref{fig:pathway_strengthening} shows that
most identified task experts exhibit positive activation-frequency
differences on five of the six tasks. Specifically, 106/157, 73/88,
53/62, 28/34, and 55/72 task experts are more frequently activated
under TEXAS than under SFT on GSM8K, MATH500, HumanEval, MBPP, and
IFEval, respectively.

At the instance level, task-expert activation gains are compared
between base-model failures that are corrected after adaptation and
those that remain incorrect. For an adapted model $m$, let $A_i^m$
denote the mean activation rate of the discovered task experts on
instance $i$, with the instance-level activation gain defined as
$\delta_i^m=A_i^m-A_i^{\mathrm{Base}}$. Among the instances that the
base model fails to solve, let $\mathcal{C}_\tau^m$ and
$\mathcal{U}_\tau^m$ denote those corrected by model $m$ and those
remaining incorrect, respectively. The correction-conditioned
activation gap is
\[
G_\tau^m
=
\mathbb{E}_{i\in\mathcal{C}_\tau^m}[\delta_i^m]
-
\mathbb{E}_{i\in\mathcal{U}_\tau^m}[\delta_i^m].
\]

The bottom panel of Figure~\ref{fig:pathway_strengthening} reports
$\Delta G_\tau =
G_\tau^{\mathrm{TEXAS}}-G_\tau^{\mathrm{SFT}}$. The difference is
significantly positive on the same five tasks, with the largest values
observed on MATH500 and IFEval. MMLU is the only exception to this
overall pattern: only 20/58 task experts show positive activation
changes, and its $\Delta G_\tau$ is significantly negative. This
mirrors the smaller performance advantage of TEXAS on MMLU in
Table~\ref{tab:main_results}. A possible explanation is that MMLU spans
heterogeneous knowledge domains, for which task-relevant computation
may be distributed across a more diffuse and less consistent set of
experts, limiting the pathway reinforcement observed under TEXAS.

The analyses above provide empirical evidence that TEXAS reinforces
task-expert pathways during adaptation. We
further argue that this reinforcement can be understood through the
gradient reweighting induced by task-expert-aware supervision. Specifically, let $\mathcal{W}_{\mathcal{B}}$ denote the answer tokens
in mini-batch $\mathcal{B}$ that are upweighted by TEXAS. Since TEXAS assigns weight $\alpha$ to these tokens and unit weight
elsewhere, for any trainable parameter $\psi$, including expert
parameters $\theta_{\ell,e}$ and router parameters $\phi_{\ell}$, we
have
\[
\nabla_{\psi}\mathcal{L}_{\mathrm{TEXAS}}
=
\nabla_{\psi}\mathcal{L}_{\mathrm{SFT}}
+
\frac{\alpha-1}{N_{\mathcal{B}}}
\sum_{(i,t)\in\mathcal{W}_{\mathcal{B}}}
\nabla_{\psi}\mathrm{CE}_{i,t},
\]
where $\mathcal{L}_{\mathrm{SFT}}$ is the unweighted
cross-entropy objective evaluated on the same mini-batch and at the
same model parameters. For expert parameters, the additional term updates only
experts activated along the corresponding token pathways. For router
parameters, it propagates through the gate weights under the current
top-$k$ assignments.

Although multiplying an individual token-level gradient by a positive
weight does not change its direction, selectively upweighting
task-expert-activating tokens changes their relative contribution to
the aggregate gradient. It can therefore alter both the magnitude and
direction of the resulting expert and router updates. By giving these
tokens greater influence during adaptation, TEXAS places greater update
emphasis on computation pathways involving the discovered task
experts, providing a plausible mechanism for the observed pathway
strengthening.

\paragraph{TEXAS concentrates supervision on task-relevant tokens.}
To assess whether task-expert activation identifies task-relevant
content, we measure rule-based enrichment on 2,048 base-model failed
training instances per task for GSM8K, MBPP, and IFEval using OLMoE,
restricting the analysis to non-special assistant answer tokens. For
each instance, we compare three size-matched token sets:
\emph{TEXAS-Weighted Tokens}, comprising answer positions assigned
weight $\alpha$ during TEXAS training; \emph{Random Tokens}, comprising
uniformly sampled answer positions; and \emph{High-Loss Tokens},
comprising positions with the largest base-model token-level
cross-entropy losses.

Task relevance is defined by task-specific rules covering mathematical
expressions and reasoning terms for GSM8K, executable code and syntax
for MBPP, and explicit constraints and structural markers for IFEval.
The complete task-relevance rules are provided in Appendix~\ref{app:analysis_details}. After pooling token positions across the sampled instances,
let $\mathcal{S}$ denote a selected token set,
$\mathcal{T}_{\mathrm{rel}}$ the task-relevant positions, and
$\mathcal{T}_{\mathrm{all}}$ all answer-token positions. We compute
\[
\operatorname{Enrich}(\mathcal{S})
=
\frac{
|\mathcal{S}\cap\mathcal{T}_{\mathrm{rel}}|/|\mathcal{S}|
}{
|\mathcal{T}_{\mathrm{rel}}|/|\mathcal{T}_{\mathrm{all}}|
}.
\]
Values greater than 1 indicate that the selected token set contains a
higher proportion of task-relevant positions than the complete
answers.

\begin{table}[t]
\centering
\small
\setlength{\tabcolsep}{5pt}
\begin{tabular}{lccc}
\toprule
\textbf{Token Set}
& \textbf{GSM8K}
& \textbf{MBPP}
& \textbf{IFEval} \\
\midrule
Random Tokens          & 0.99 & 1.00 & 1.00 \\
High-Loss Tokens       & 1.13 & 0.72 & 0.45 \\
\textbf{TEXAS-Weighted Tokens}
                       & \textbf{2.21}
                       & \textbf{1.35}
                       & \textbf{2.22} \\
\bottomrule
\end{tabular}
\caption{
Rule-based task-relevance enrichment of size-matched token sets on
OLMoE. Values greater than 1 indicate
enrichment relative to all answer tokens.
}
\label{tab:token_relevance}
\end{table}

As shown in Table~\ref{tab:token_relevance}, random tokens obtain
enrichment ratios close to 1. High-loss tokens show inconsistent task
relevance: they are only slightly enriched on GSM8K and fall below the
answer-level baseline on MBPP and IFEval. In
contrast, TEXAS-weighted tokens achieve enrichment ratios of 2.21,
1.35, and 2.22 on GSM8K, MBPP, and IFEval, respectively, outperforming
both size-matched comparison sets on every task. These results indicate
that task-expert activation provides a more targeted signal than token
loss for allocating stronger supervision, complementing the
TEXAS-AllTok ablation by explaining the benefit of token-level
selectivity.

\section{Conclusion}

This work highlights a new way to exploit expert specialization in
MoE adaptation: task-expert activations can serve not only as signals
for expert selection or routing optimization, but also as signals for
deciding where training supervision should be concentrated. Based on
this perspective, we introduced Task-Expert-Aware Supervision
(TEXAS), which combines correctness-conditioned task expert discovery
with token-level supervision allocation. Experiments across diverse
models and tasks, together with controlled ablations, expert masking,
and token-level analyses, support the effectiveness of both components.
More broadly, TEXAS suggests that internal computation pathways can
provide useful guidance for allocating learning signals during
adaptation. Future work could extend this direction by using richer
outcome signals beyond binary correctness and dynamically updating task
experts as the model evolves during multitask or continual adaptation.

\bibliography{aaai2027}

\appendix

\setcounter{secnumdepth}{2}
\setcounter{section}{0}
\setcounter{subsection}{0}
\renewcommand{\thesection}{\Alph{section}}
\renewcommand{\thesubsection}{\thesection.\arabic{subsection}}

\appendix

\section{Additional Method and Discovery Details}
\label{app:method_details}

\subsection{Correctness Partition Construction}
\label{app:success_criteria}

For each model--task setting, we obtain a prediction from the
unadapted base model for every instance in the paired training set.
Instances satisfying the task-specific success criterion form
$\Dplus$, while all remaining instances form $\Dminus$. Generation
failures, unparsable outputs, compilation or execution errors, and
timeouts are treated as unsuccessful where applicable.

All correctness partitions are constructed once using greedy decoding
and are reused during expert discovery and downstream fine-tuning. Table~\ref{tab:success_criteria}
summarizes the paired training data and success criteria.

\begin{table*}[t]
\centering
\small
\setlength{\tabcolsep}{5pt}
\begin{tabularx}{\textwidth}{l l X}
\toprule
Target benchmark
& Discovery/training data
& Partition procedure and success criterion \\
\midrule

GSM8K
& MetaMathQA, GSM8K-derived subset
& Four-shot mathematical-reasoning generation; exact match after
numeric answer extraction and normalization. \\

MATH500
& MetaMathQA, MATH-derived subset
& Four-shot mathematical-reasoning generation; normalized exact match
of the extracted final boxed answer. \\

HumanEval
& CodeAlpaca data with executable tests
& Code generation followed by execution of the associated
instance-level tests; all tests must pass. \\

MBPP
& MBPP-like OpenCodeInstruct data
& Python function generation followed by execution of the associated
tests; all tests must pass. \\

MMLU
& MMLU auxiliary training set
& Free-form generation of an answer option from A/B/C/D; the extracted
option must equal the gold answer. \\

IFEval
& RECAST-30K
& Instruction-following generation; all associated constraints must
pass their corresponding loose checkers. \\

\bottomrule
\end{tabularx}
\caption{
Training-set procedures used to construct $\Dplus$ and $\Dminus$.
Task names denote the target downstream benchmarks, while partitioning
is performed on their paired training sets.
}
\label{tab:success_criteria}
\end{table*}

\subsection{Correctness-Conditioned Expert Discovery}
\label{app:discovery_details}

Task experts are discovered exclusively from the training set paired
with each target benchmark. We consider only routed MoE experts, excluding
shared or dense experts.

For each training instance $i$, we perform a teacher-forced forward
pass over the reference answer and collect the model's native top-$k$
routing assignments on valid assistant answer tokens. Let
$\mathcal{A}_i$ denote these positions and let
$R_{i,t}^{(\ell)}$ denote the selected experts at token $t$ and layer
$\ell$. The instance-level activation rate of expert $e$ is

\begin{equation}
a_{i,\ell,e}
=
\frac{1}{|\mathcal{A}_i|}
\sum_{t\in\mathcal{A}_i}
\mathbf{1}
\left[
e\in R_{i,t}^{(\ell)}
\right].
\end{equation}

Activation is defined by binary membership in the native top-$k$
routing set rather than by gate-weight magnitude. For every
layer--expert pair, we compare the activation rates on $\Dplus$ and
$\Dminus$ using a one-sided Welch's $t$-test with alternative
$\mu_{\ell,e}^{+}>\mu_{\ell,e}^{-}$. Benjamini--Hochberg correction is
applied over all layer--expert tests within each model--task setting,
and experts with adjusted $p<0.05$ are treated as significant.

To exclude significant but rarely activated experts, we additionally
retain the top-$K$ experts in each layer according to their mean
activation rates on successful instances. The final task-expert set is

\begin{equation}
\mathcal{E}_{\mathrm{task}}^{(\ell)}
=
\mathcal{E}_{\mathrm{sig}}^{(\ell)}
\cap
\mathcal{E}_{\mathrm{cand}}^{(\ell)}.
\end{equation}

Unless otherwise specified, $K=2k$, where $k$ is the model's native
number of routed experts per token.
Table~\ref{tab:discovery_stats} summarizes the resulting correctness
partitions and the number of retained task experts for each
model--task setting.

\begin{table}[t]
\centering
\small
\begin{tabular}{lrrrr}
\toprule
Task
& $|\Dplus|$
& $|\Dminus|$
& Success
& Experts \\
\midrule

\multicolumn{5}{l}{\textbf{\deepseek}} \\
GSM8K
& 82{,}854 & 156{,}992 & 34.5\% & 154 \\
MATH500
& 27{,}234 & 66{,}053 & 29.2\% & 162 \\
HumanEval
& 3{,}082 & 4{,}000 & 43.5\% & 96 \\
MBPP
& 13{,}321 & 106{,}679 & 11.1\% & 149 \\
MMLU
& 28{,}722 & 71{,}120 & 28.8\% & 170 \\
IFEval
& 1{,}743 & 10{,}568 & 14.2\% & 56 \\

\midrule
\multicolumn{5}{l}{\textbf{\olmoe}} \\
GSM8K
& 26{,}739 & 213{,}107 & 11.1\% & 157 \\
MATH500
& 2{,}909 & 90{,}378 & 3.1\% & 88 \\
HumanEval
& 2{,}826 & 4{,}256 & 39.9\% & 62 \\
MBPP
& 10{,}302 & 109{,}698 & 8.6\% & 34 \\
MMLU
& 53{,}274 & 46{,}568 & 53.4\% & 58 \\
IFEval
& 722 & 11{,}589 & 5.9\% & 72 \\

\midrule
\multicolumn{5}{l}{\textbf{\qwen}} \\
GSM8K
& 150{,}863 & 88{,}983 & 62.9\% & 75 \\
MATH500
& 48{,}229 & 45{,}058 & 51.7\% & 87 \\
HumanEval
& 3{,}300 & 3{,}782 & 46.6\% & 52 \\
MBPP
& 42{,}840 & 77{,}160 & 35.7\% & 42 \\
MMLU
& 63{,}000 & 36{,}842 & 63.1\% & 36 \\
IFEval
& 3{,}519 & 8{,}792 & 28.6\% & 46 \\

\bottomrule
\end{tabular}
\caption{
Training-set correctness partitions and retained task-expert counts.
Success denotes the proportion of training instances in $\Dplus$.
Task names refer to the target benchmarks, while discovery is performed
on their paired training sets.
}
\label{tab:discovery_stats}
\end{table}

\subsection{Training-Time Token Weighting}
\label{app:weighting_details}

The discovered task-expert masks remain fixed throughout fine-tuning,
while token-level task-expert hits are recomputed from the current
model's native top-$k$ routing assignments at every forward pass.

For an answer token $t$ in instance $i$, TEXAS assigns

\begin{equation}
w_{i,t}
=
\begin{cases}
\alpha,
&
i\in\Dminus
\ \land\
\exists\ell:
R_{i,t}^{(\ell)}
\cap
\mathcal{E}_{\mathrm{task}}^{(\ell)}
\neq\varnothing,\\
1,
& \text{otherwise}.
\end{cases}
\end{equation}

A token receives weight $\alpha$ only once, even if it activates
multiple task experts or activates task experts in multiple layers.
The default setting is $\alpha=1.2$. All other answer tokens retain
unit weight, and the loss is normalized over valid assistant answer
tokens.

Because the routing decisions are obtained from the current model,
the upweighted token positions may change during fine-tuning even
though the correctness partitions and task-expert masks remain fixed.

\section{Complete Experimental Setup}
\label{app:experimental_setup}

\subsection{Models and Adaptation Setup}
\label{app:models_backbone}

We evaluate \deepseek, \olmoe, and \qwen.
Their principal MoE configurations are summarized in
Table~\ref{tab:model_configs}.

\begin{table}[t]
\centering
\small
\begin{tabular}{lrrrr}
\toprule
Model
& Layers
& Experts/layer
& Shared
& Top-$k$ \\
\midrule
\deepseek & 27 & 64 & 2 & 6 \\
\olmoe    & 16 & 64 & 0 & 8 \\
\qwen     & 24 & 60 & 0 & 4 \\
\bottomrule
\end{tabular}
\caption{
MoE routing configurations. Shared experts are excluded from
task-expert discovery and expert selection.
}
\label{tab:model_configs}
\end{table}

All methods use LoRA adapters on the up- and down-projection matrices
of routed experts, with rank 16, scaling factor 32, and dropout 0.05.
Router parameters remain trainable, while the remaining pretrained
parameters are frozen. SFT, RoMA, and TEXAS adapt all routed experts;
ESFT adapts only its selected experts.

DeepSeek and Qwen use their model-provided tokenizers and chat
templates. For OLMoE, we use the GPT-NeoX tokenizer shipped with its checkpoint together with a
Tulu-style chat template because the checkpoint does not provide a built-in chat template.

\subsection{Training Data and Preprocessing}
\label{app:data_details}

Each benchmark is paired with a task-relevant supervised training set,
as summarized in Table~\ref{tab:data_configs}.

\begin{table}[t]
\centering
\small
\setlength{\tabcolsep}{3.5pt}
\begin{tabular}{llrr}
\toprule
Task
& Training data
& Examples
& Max length \\
\midrule
GSM
& MetaMathQA, GSM-derived
& 239{,}846
& 1{,}024 \\

MATH
& MetaMathQA, MATH-derived
& 93{,}287
& 1{,}024 \\

HE
& CodeAlpaca data with tests
& 7{,}082
& 1{,}024 \\

MBPP
& OpenCodeInstruct, MBPP-like
& 120{,}000
& 1{,}024 \\

MMLU
& MMLU auxiliary training split
& 99{,}842
& 512 \\

IF
& RECAST-30K rule subset
& 12{,}311
& 1{,}024 \\
\bottomrule
\end{tabular}
\caption{
Training datasets and sequence lengths.
}
\label{tab:data_configs}
\end{table}

The HumanEval training set contains code-generation
instances with executable tests. The MBPP training subset contains
Python programming instances and is decontaminated against the
evaluation benchmark. The IFEval training subset contains instances
with executable instruction-following constraints.

All examples are represented as user--assistant conversations and
tokenized using the corresponding chat template. Sequences are
truncated from the right to the maximum lengths in
Table~\ref{tab:data_configs}. Cross-entropy loss is computed only on
assistant answer tokens; user and formatting spans are masked.
Dynamic padding is used, and sequence packing is disabled.

\subsection{Training Configuration}
\label{app:training_hparams}

Table~\ref{tab:training_hparams} summarizes the common fine-tuning
hyperparameters. Weight decay is not applied to bias or normalization
parameters. Each run uses four GPUs, with effective batch sizes of 64
for \deepseek, 256 for \olmoe, and 128 for \qwen.

\begin{table}[t]
\centering
\small
\setlength{\tabcolsep}{5pt}
\begin{tabular}{ll}
\toprule
Hyperparameter & Value \\
\midrule
Optimizer & AdamW \\
Learning rate & $2\times10^{-5}$ \\
Scheduler & Cosine \\
Warmup ratio & 0.03 \\
Weight decay & 0.01 \\
Gradient clipping & 1.0 \\
Precision & bfloat16 \\
Training duration & One epoch \\
Training seed & 42 \\
\bottomrule
\end{tabular}
\caption{
Common downstream fine-tuning configuration. The same learning rate is
used for expert LoRA and router parameters.
}
\label{tab:training_hparams}
\end{table}

\subsection{Baseline Implementations}
\label{app:baseline_details}

\paragraph{SFT.}
SFT minimizes the standard assistant-token cross-entropy objective
using the all-expert LoRA configuration. It is equivalent to the TEXAS
objective with $\alpha=1$.

\paragraph{ESFT.}
ESFT selects routed experts independently in each MoE layer using
aggregate token-routing statistics. We use the ESFT-Token variant with
the cumulative-threshold parameter $p=0.2$. Only LoRA adapters attached
to the selected experts are trained, while the router remains
trainable.

\paragraph{RoMA.}
We first evaluated the router-only configuration used in the original
RoMA formulation. Under our downstream adaptation setting, however,
router-only RoMA performed substantially worse than standard SFT; on
OLMoE--GSM8K, it was 15.6 points below SFT. We therefore use the same
all-expert LoRA backbone as SFT and TEXAS, with both routed-expert
adapters and router parameters trainable, and add the RoMA
routing-alignment objective on top of this common backbone.

For each training instance, RoMA retrieves three semantically similar
examples that are successfully solved by the base model. Retrieval uses
normalized \texttt{sentence-transformers/all-MiniLM-L6-v2} embeddings
and cosine similarity. The retrieved examples' cached last-position
routing profiles are combined into a similarity-weighted target, and
RoMA minimizes the mean squared error between the current and target
routing distributions. We use $\sigma=0.4$ and
$\lambda_{\mathrm{RoMA}}=1.0$.

\paragraph{TEXAS.}
TEXAS uses the correctness-conditioned task experts described in
Appendix~\ref{app:method_details}. The default candidate size is
$K=2k$, and the default token weight is $\alpha=1.2$. The discovered
expert masks remain fixed, while token-level task-expert hits are
recomputed from the current routing assignments during fine-tuning.

\subsection{Evaluation Settings}
\label{app:evaluation_details}

All benchmark results are obtained through response generation rather
than fixed-answer logit scoring. Each checkpoint is evaluated with
temperature 0.2, top-$p=1.0$, and generation seeds 42, 2026, and 330.
Table~\ref{tab:eval_configs} summarizes the benchmark sizes, generation
lengths, metrics, and evaluation criteria. GSM8K and MATH500 use
normalized exact match after answer extraction; HumanEval and MBPP
execute generated code against their associated tests with a 10-second
timeout; MMLU extracts an answer option from A/B/C/D; and IFEval reports
prompt-level loose instruction-following accuracy.

\begin{table}[t]
\centering
\scriptsize
\setlength{\tabcolsep}{3pt}
\begin{tabular}{lrrrl}
\toprule
Task
& Examples
& Max tokens
& Metric
& Evaluation \\
\midrule
GSM8K
& 1{,}319
& 512
& Acc.
& Numeric exact match \\

MATH500
& 500
& 1{,}024
& Acc.
& Final-answer exact match \\

HumanEval
& 164
& 512
& Pass@1
& Official unit tests \\

MBPP
& 500
& 512
& Pass@1
& Benchmark unit tests \\

MMLU
& 14{,}042
& 8
& Acc.
& Generated A/B/C/D option \\

IFEval
& 541
& 1{,}280
& Loose acc.
& Instruction checkers \\
\bottomrule
\end{tabular}
\caption{
Benchmark evaluation settings. All results use stochastic generation
with temperature 0.2, top-$p=1.0$, and seeds 42, 2026, and 330.
}
\label{tab:eval_configs}
\end{table}

\subsection{Computational Environment}
\label{app:infrastructure}

Experiments are conducted on a server with eight NVIDIA RTX 5880 Ada
GPUs, each with approximately 49\,GB of memory, and two AMD EPYC 9554
CPUs. Individual fine-tuning runs typically use four GPUs.
Table~\ref{tab:infra} summarizes the principal software environment.
Training uses DeepSpeed ZeRO Stage 2, generation-based evaluation uses
vLLM, and FlashAttention is enabled where supported.

\begin{table}[t]
\centering
\small
\setlength{\tabcolsep}{5pt}
\begin{tabular}{ll}
\toprule
Component & Version \\
\midrule
CUDA / driver & 12.8 / 580.82.07 \\
Python & 3.12.12 \\
PyTorch & 2.10.0+cu128 \\
Transformers & 4.57.6 \\
PEFT & 0.13.2 \\
DeepSpeed & 0.17.2, ZeRO-2 \\
vLLM & 0.17.1 \\
\bottomrule
\end{tabular}
\caption{
Principal software environment used for training and evaluation.
}
\label{tab:infra}
\end{table}

\section{Ablation and Robustness Details}
\label{app:ablation_robustness}

\subsection{Ablation Variants}
\label{app:ablation_definitions}

We conduct ablation experiments on OLMoE using GSM8K, MBPP, and
IFEval. Let $\mathcal{E}_{\mathrm{CC}}^{(\ell)}$ denote the
correctness-conditioned task experts in layer $\ell$.

\paragraph{TEXAS-Freq.}
TEXAS-Freq replaces the correctness-conditioned expert mask with a
frequency-based mask while retaining the token-level weighting rule of TEXAS. Aggregate frequency is
computed from cached last-position routing records: for each training
instance and layer, the native top-$k$ selected experts are converted
into binary indicators and averaged across instances. In every layer,
the highest-frequency experts are selected to match the cardinality of
$\mathcal{E}_{\mathrm{CC}}^{(\ell)}$. The statistic therefore uses
binary expert selections rather than gate-weight magnitudes.

\paragraph{TEXAS-Route.}
TEXAS-Route uses the same correctness-conditioned task-expert masks as TEXAS, but changes how the
discovered task experts are used. Instead of using their activation as
a signal for cross-entropy upweighting, TEXAS-Route directly encourages
the router to assign greater probability mass to them.

Let $p_{i,t}^{(\ell)}(e)$ denote the current post-softmax router
probability assigned to expert $e$ at answer token $t$ in layer
$\ell$. As in TEXAS, the relevant positions in each layer are

\begin{equation}
\mathcal{S}_{\ell}
=
\left\{
(i,t):
i\in\Dminus,\;
t\in\mathcal{A}_i,\;
R_{i,t}^{(\ell)}
\cap
\mathcal{E}_{\mathrm{CC}}^{(\ell)}
\neq\varnothing
\right\},
\end{equation}

where $R_{i,t}^{(\ell)}$ is the model's native top-$k$ routed-expert
set. At each selected position, the routing mass assigned to the same
task-expert set used by TEXAS is

\begin{equation}
m_{i,t}^{(\ell)}
=
\sum_{e\in\mathcal{E}_{\mathrm{CC}}^{(\ell)}}
p_{i,t}^{(\ell)}(e).
\end{equation}

The layer-wise routing loss is

\begin{equation}
\mathcal{L}_{\mathrm{route}}^{(\ell)}
=
-\frac{1}{|\mathcal{S}_{\ell}|}
\sum_{(i,t)\in\mathcal{S}_{\ell}}
\log\!\left(
m_{i,t}^{(\ell)}+10^{-8}
\right).
\end{equation}

Let
$\mathcal{J}_{\mathrm{act}}
=
\{\ell:|\mathcal{S}_{\ell}|>0\}$
denote the layers containing at least one selected position. The
overall routing loss is

\begin{equation}
\mathcal{L}_{\mathrm{route}}
=
\frac{1}{|\mathcal{J}_{\mathrm{act}}|}
\sum_{\ell\in\mathcal{J}_{\mathrm{act}}}
\mathcal{L}_{\mathrm{route}}^{(\ell)}.
\end{equation}

TEXAS-Route optimizes

\begin{equation}
\mathcal{L}_{\mathrm{TEXAS\text{-}Route}}
=
\mathcal{L}_{\mathrm{SFT}}
+
\lambda_{\mathrm{route}}
\mathcal{L}_{\mathrm{route}},
\end{equation}

where $\lambda_{\mathrm{route}}=0.1$.

\paragraph{TEXAS-AllInst.}
TEXAS-AllInst uses the same correctness-conditioned expert masks as TEXAS but applies task-expert-aware
weighting to all training instances rather than only to $\Dminus$.

\paragraph{TEXAS-AllTok.}
TEXAS-AllTok applies task-expert-aware
weighting to $\Dminus$ but assigns weight
$\alpha$ to every valid answer token, irrespective of
whether the token activates a discovered task expert. This variant
tests whether uniformly emphasizing failed instances can reproduce the
benefit of task-expert-aware token selection.

\subsection{Ablation Protocol}
\label{app:ablation_protocol}

All ablation variants use the same OLMoE backbone, supervised training
data, LoRA configuration, trainable router setup, and optimization
settings as the corresponding main
experiments. One checkpoint is trained for each task and variant using seed 42.
Reported means and standard deviations are computed from stochastic
test-time generation with temperature $0.2$, top-$p=1.0$, and seeds
42, 2026, and 330.

\subsection{Hyperparameter Robustness}
\label{app:robustness_details}

We evaluate the full Cartesian product of candidate-filtering sizes
and token-level loss weights

\begin{equation}
\begin{aligned}
K &\in
\left\{
k,\,
1.5k,\,
2k,\,
3k,\,
\text{No filt.}
\right\},\\
\alpha &\in
\left\{
1.05,\,
1.1,\,
1.2,\,
1.5,\,
2.0
\right\}.
\end{aligned}
\end{equation}

Each configuration is fine-tuned separately on GSM8K, MBPP, and
IFEval using OLMoE. For finite $K$, experts are ranked within each
layer by their mean activation rate on successful training instances,
and the top-$K$ candidates are intersected with the statistically
significant experts. OLMoE uses native $k=8$, so $1.5k$ corresponds to
12 candidates per layer. The ``No filt.'' setting retains all experts
passing the corrected significance threshold.

Let
\(
\mathcal{T}_{\mathrm{rob}}
=
\{\mathrm{GSM8K},\mathrm{MBPP},\mathrm{IFEval}\}
\).
Each heatmap cell reports the average absolute improvement over the
corresponding SFT baselines:

\begin{equation}
\Delta(K,\alpha)
=
\frac{1}{|\mathcal{T}_{\mathrm{rob}}|}
\sum_{\tau\in\mathcal{T}_{\mathrm{rob}}}
\left[
M_{\tau}(K,\alpha)
-
M_{\tau}^{\mathrm{SFT}}
\right].
\end{equation}

All robustness configurations use the same data, training seed,
optimization settings, and evaluation protocol; only $K$ and
$\alpha$ are varied.

\FloatBarrier
\section{Analysis Protocols and Additional Statistics}
\label{app:analysis_details}

All analyses use \olmoe~and the task-expert sets discovered from the
corresponding training data. These sets remain fixed throughout
evaluation. Unless otherwise stated, routing is collected by
teacher-forcing each model on the evaluation prompt concatenated with
its generated response and recording native top-$k$ routing decisions
on valid answer tokens.

\subsection{SuccessGap Analysis}
\label{app:successgap}

For evaluation instance $i$, the activation rate of expert $e$ in
layer $\ell$ is

\begin{equation}
a_{i,\ell,e}
=
\frac{1}{|\mathcal{A}_i|}
\sum_{t\in\mathcal{A}_i}
\mathbf{1}
\left[
e\in R_{i,t}^{(\ell)}
\right],
\end{equation}

where $\mathcal{A}_i$ denotes the valid generated-answer positions.
For task-expert set $\mathcal{E}_{\tau}$, we compute

\begin{equation}
\operatorname{SuccessGap}(\mathcal{E}_{\tau})
=
\frac{1}{|\mathcal{E}_{\tau}|}
\sum_{(\ell,e)\in\mathcal{E}_{\tau}}
\left(
\mu_{\ell,e}^{+}
-
\mu_{\ell,e}^{-}
\right),
\end{equation}

where $\mu_{\ell,e}^{+}$ and $\mu_{\ell,e}^{-}$ are the mean activation
rates on successfully and unsuccessfully solved evaluation instances,
respectively.

For the ESFT comparison, we use the aggregate token-routing frequency
criterion underlying ESFT-Token. To isolate expert-selection quality
from differences in selection size, we select the highest-frequency
experts in each layer while matching the layer-wise cardinality of the
correctness-conditioned expert set. The random baseline uses
1,000 independently sampled expert sets with the same layer-wise
cardinalities. The resulting SuccessGap comparisons are reported in
the main paper.

\subsection{Expert Masking}
\label{app:masking_protocol}

We assess the functional importance of selected experts by setting
their down-projection outputs to zero in the SFT-adapted \olmoe~checkpoint.
Routing decisions and all remaining parameters are unchanged. For task
$\tau$, the common masking budget in layer $\ell$ is

\begin{equation}
b_{\tau}^{(\ell)}
=
\min
\left(
2,\,
|\mathcal{E}_{\mathrm{CC},\tau}^{(\ell)}|
\right).
\end{equation}

The same layer-wise budget is used for random, ESFT-based, and
correctness-conditioned masking. For each condition, three mask sets
are sampled using seeds 42, 2026, and 330. Table~\ref{tab:masking_results}
reports the resulting performance.

\begin{table}[t]
\centering
\small
\setlength{\tabcolsep}{3.5pt}
\begin{tabular}{lrrrr}
\toprule
Task
& Unmasked
& Random
& ESFT
& \textbf{CC} \\
\midrule
GSM8K
& 30.8
& $25.8{\pm}1.8$
& $12.1{\pm}3.1$
& \textbf{$7.0{\pm}0.9$} \\

MATH500
& 4.9
& $4.5{\pm}0.7$
& $4.1{\pm}0.7$
& \textbf{$2.9{\pm}0.1$} \\

HumanEval
& 13.8
& $13.4{\pm}0.8$
& $11.5{\pm}0.7$
& \textbf{$9.2{\pm}0.5$} \\

MBPP
& 13.8
& $10.4{\pm}4.0$
& $11.3{\pm}1.9$
& $7.6{\pm}1.4$ \\

MMLU
& 51.0
& $48.9{\pm}1.6$
& $48.1{\pm}0.4$
& \textbf{$46.4{\pm}0.7$} \\

IFEval
& 26.9
& $25.6{\pm}1.1$
& $23.4{\pm}1.1$
& \textbf{$21.6{\pm}0.4$} \\
\bottomrule
\end{tabular}
\caption{
OLMoE performance under layer-wise cardinality-matched expert
masking. Unmasked denotes the SFT-adapted checkpoint without expert
masking. Values report means and standard deviations over three
sampled masks. CC denotes correctness-conditioned experts.
}
\label{tab:masking_results}
\end{table}

\subsection{Task-Expert Pathway Analysis}
\label{app:pathway_analysis}

Both pathway analyses use the seed-42 SFT and TEXAS checkpoints and
their corresponding seed-42 evaluation outputs.

\paragraph{Expert-level activation changes.}
For each discovered task expert $(\ell,e)$, we compute

\begin{equation}
\Delta_{\ell,e}^{\mathrm{act}}
=
\mu_{\ell,e}^{\mathrm{TEXAS}}
-
\mu_{\ell,e}^{\mathrm{SFT}},
\end{equation}

where $\mu_{\ell,e}^{m}$ is its mean generated-answer activation rate
under model $m$. Positive values indicate that TEXAS activates the
task expert more frequently than SFT. The expert-level results are
reported in the main paper.

\paragraph{Correction-conditioned activation gap.}
For adapted model $m$, let $A_i^m$ denote the mean activation rate of
the discovered task experts on instance $i$. Relative to the base
model, define

\begin{equation}
\delta_i^m
=
A_i^m
-
A_i^{\mathrm{Base}}.
\end{equation}

Among instances failed by the base model, let
$\mathcal{C}_{\tau}^{m}$ contain those corrected by model $m$ and
$\mathcal{U}_{\tau}^{m}$ those remaining incorrect. We compute

\begin{equation}
G_{\tau}^{m}
=
\mathbb{E}_{i\in\mathcal{C}_{\tau}^{m}}
[\delta_i^m]
-
\mathbb{E}_{i\in\mathcal{U}_{\tau}^{m}}
[\delta_i^m],
\end{equation}

and compare TEXAS with SFT through

\begin{equation}
\Delta G_{\tau}
=
G_{\tau}^{\mathrm{TEXAS}}
-
G_{\tau}^{\mathrm{SFT}}.
\end{equation}

A positive $\Delta G_{\tau}$ indicates that, relative to SFT, TEXAS
concentrates larger task-expert activation gains on base-model failures
that become correct. We compute 95\% percentile bootstrap intervals
using 10,000 resamples of the base-model-failed instances with seed 42.

\subsection{Task-Relevance Enrichment}
\label{app:token_relevance}

We analyze GSM8K, MBPP, and IFEval, representing mathematical
reasoning, code generation, and instruction following, respectively.
For each task, we use the first 2,048 base-model-failed training
instances containing valid assistant-answer tokens. Special tokens and
non-assistant positions are excluded.

\paragraph{Size-matched token sets.}
Let $\mathcal{A}_i$ denote the valid answer-token positions of instance
$i$. The TEXAS-weighted token set is

\begin{equation}
\mathcal{S}_{i}^{\mathrm{TEXAS}}
=
\left\{
t\in\mathcal{A}_i:
\exists\ell,\;
R_{i,t}^{(\ell)}
\cap
\mathcal{E}_{\mathrm{CC}}^{(\ell)}
\neq\varnothing
\right\}.
\end{equation}

For each instance, we construct two comparison sets with the same
cardinality as $\mathcal{S}_{i}^{\mathrm{TEXAS}}$. Random Tokens are
sampled uniformly without replacement from $\mathcal{A}_i$ using seed
42. High-Loss Tokens are the answer positions with the largest
base-model token-level cross-entropy losses. The same instances are
used for all three token sets.

\paragraph{Task-relevance rules.}
Task relevance is determined directly from decoded answer-token
strings rather than from character spans in the fully detokenized
answer. Table~\ref{tab:relevance_rules} summarizes the task-specific
token classes, and Table~\ref{tab:relevance_examples} provides
representative substrings that trigger these rules.

\begin{table}[t]
\centering
\scriptsize
\setlength{\tabcolsep}{3pt}
\begin{tabularx}{\columnwidth}{
    >{\raggedright\arraybackslash}p{0.85cm}
    >{\raggedright\arraybackslash}X}
\toprule
Task & Task-relevant token classes \\
\midrule

GSM8K
&
Numbers, arithmetic and comparison symbols, units, currency and
percentage symbols, answer markers, and explicit reasoning terms. \\

MBPP
&
Python identifiers, keywords, operators, delimiters, brackets,
punctuation, and code-structure tokens; comments and docstrings are
excluded. \\

IFEval
&
Constraint terms, counts, lexical and formatting requirements,
section markers, markup, and relevant punctuation. \\

\bottomrule
\end{tabularx}
\caption{
Token classes used by the rule-based task-relevance analysis.
}
\label{tab:relevance_rules}
\end{table}

\begin{table}[t]
\centering
\scriptsize
\setlength{\tabcolsep}{3pt}
\begin{tabularx}{\columnwidth}{
    >{\raggedright\arraybackslash}p{0.75cm}
    >{\raggedright\arraybackslash}X
    >{\raggedright\arraybackslash}p{2.45cm}}
\toprule
Task & Answer excerpt & Matched token types \\
\midrule

GSM8K
&
\texttt{\$25 + \$15.20 + \$6.80 = \$47;}
\newline
\texttt{16 * \$47 = \$752; \#\#\#\# 752}
&
Numbers, currency symbols, arithmetic operators, equality signs, and
final-answer markers. \\

MBPP
&
\texttt{def is\_balanced\_}
\newline
\texttt{brackets(expression):}
&
Function names, identifiers, Python keywords, parentheses, underscores,
and punctuation delimiters. \\

IFEval
&
\texttt{\# How to Choose ...}
\newline
\texttt{- **Uptime**: ... **99.9\%** ...}
\newline
\texttt{- **Scalability**: ...}
&
Markdown heading and bullet markers, bold delimiters, punctuation,
digits, and percentage symbols. \\

\bottomrule
\end{tabularx}
\caption{
Illustrative rule matches from base-model-failed training instances.
The excerpts show decoded answer-token substrings that trigger the
task-relevance rules.
}
\label{tab:relevance_examples}
\end{table}

\paragraph{Enrichment metric.}
After pooling token positions across the selected instances, let
$\mathcal{S}$ denote one of the three token sets,
$\mathcal{T}_{\mathrm{rel}}$ the task-relevant positions, and
$\mathcal{T}_{\mathrm{all}}$ all valid answer-token positions. We
compute

\begin{equation}
\operatorname{Enrich}(\mathcal{S})
=
\frac{
|\mathcal{S}\cap\mathcal{T}_{\mathrm{rel}}|/|\mathcal{S}|
}{
|\mathcal{T}_{\mathrm{rel}}|/|\mathcal{T}_{\mathrm{all}}|
}.
\end{equation}

Values above 1 indicate that the selected set contains a higher
proportion of task-relevant tokens than the complete assistant
answers.

\FloatBarrier
\section{Computational Cost and Efficiency}
\label{app:compute}

TEXAS introduces a one-time task-expert discovery stage before
fine-tuning. We report this offline cost separately from fine-tuning
cost because the discovered task experts can be cached and reused
across subsequent runs. During fine-tuning, TEXAS uses the same number
of model forward and backward passes as SFT.

\subsection{Offline Discovery Cost}
\label{app:offline_cost}

Offline discovery consists of base-model generation and evaluation on
the training data, teacher-forced routing collection over reference
answers, and statistical expert selection. The first two stages require
forward passes only. They are mutually independent because routing
collection uses reference answers rather than model-generated
responses, and can therefore be executed concurrently when separate
resources are available. Statistical expert selection is performed
after both stages and has negligible cost.

Table~\ref{tab:offline_cost} reports representative measurements for
OLMoE--GSM8K. Each forward-only stage uses four GPUs, and the total
shown in the table corresponds to sequential execution.

\begin{table}[t]
\centering
\small
\setlength{\tabcolsep}{5pt}
\begin{tabular}{lr}
\toprule
Stage & Wall time \\
\midrule
Base-model inference & 0.3 h \\
Routing collection & 1.2 h \\
Statistical selection & $<1$ min \\
\midrule
Sequential total & 1.5 h \\
\bottomrule
\end{tabular}
\caption{
Offline discovery cost for OLMoE--GSM8K using 4 X
NVIDIA RTX 5880 Ada GPUs. Sequential execution
takes 1.5 hours, corresponding to approximately 6 GPU hours. The two
forward-only stages are independent and can be executed concurrently
when separate resources are available.
}
\label{tab:offline_cost}
\end{table}

The sequential discovery time is approximately 26\% of a single TEXAS
fine-tuning run. Because the base-model predictions, correctness
partitions, reference-answer routing records, and discovered
task-expert sets are cached, this cost is incurred only once per
model--task setting and can be amortized across subsequent fine-tuning
runs.

\subsection{Fine-Tuning Efficiency}
\label{app:training_cost}

Table~\ref{tab:training_cost} compares representative OLMoE--GSM8K
fine-tuning costs under matched settings: 4 GPUs, 938 training
steps, and an effective batch size of 256. The table reports
fine-tuning time only and excludes method-specific offline
preprocessing.

\begin{table}[t]
\centering
\small
\setlength{\tabcolsep}{3.5pt}
\begin{tabular}{lrrrr}
\toprule
Method
& Time
& Relative
& Memory
& Trainable \\
&
(h)
&
time
&
(GB/GPU)
&
params. (M) \\
\midrule
SFT
& 5.9
& $1.00\times$
& 43.2
& 100.7 \\

ESFT
& 5.8
& $0.98\times$
& 41.6
& 12.6 \\

RoMA
& 8.5
& $1.44\times$
& 43.5
& 100.7 \\

TEXAS
& 5.8
& $0.98\times$
& 43.2
& 100.7 \\
\bottomrule
\end{tabular}
\caption{
Representative OLMoE--GSM8K fine-tuning cost under matched training
settings.
}
\label{tab:training_cost}
\end{table}

SFT provides the fine-tuning-time reference and requires no
method-specific offline preparation. ESFT trains fewer parameters and
uses slightly less memory because it attaches LoRA adapters only to
selected routed experts. However, it does not reduce the number of
experts activated during the forward pass, so its wall-clock time
remains similar to SFT.

RoMA retains the same trainable parameter count and number of model
passes as SFT, but constructing and applying neighbor-based routing
targets introduces additional training-time operations, increasing its
wall-clock cost. TEXAS also retains the same trainable parameters and
model passes as SFT, while adding only token-mask construction and
loss reweighting. Its fine-tuning time and memory usage therefore
remain on par with SFT.

Like TEXAS, ESFT and RoMA require method-specific one-time
preprocessing to construct expert-selection statistics or
neighbor-based routing targets. Thus, offline preparation is not
unique to TEXAS among the specialized adaptation methods. Relative to
SFT, TEXAS adds approximately 1.5 hours of reusable discovery cost in
this representative setting, while preserving SFT-level fine-tuning
efficiency and providing the consistent performance improvements
reported in the main experiments.

\end{document}